\documentclass[letterpaper, 10 pt, conference]{ieeeconf}
\IEEEoverridecommandlockouts    % This command is only needed if
\usepackage{graphics}           
\usepackage{times}              
\usepackage{amsmath}            
\usepackage{amssymb}            
\usepackage{graphicx}
\usepackage{algorithm}
\usepackage[noend]{algpseudocode}
\usepackage{booktabs}
\usepackage{color}
\definecolor{instructioncolor}{rgb}{.5,.5,.5}

\usepackage[font=small]{caption}

\def\figref#1{Fig.~\ref{#1}}
\def\tabref#1{Tab.~\ref{#1}}
\def\eqref#1{Eq.~(\ref{#1})}

\makeatletter
\usepackage{xspace}
\DeclareRobustCommand\onedot{\futurelet\@let@token\@onedot}
\def\@onedot{\ifx\@let@token.\else.\null\fi\xspace}

\makeatother

\usepackage{array}
\newcolumntype{L}[1]{>{\raggedright\let\newline\\\arraybackslash\hspace{0pt}}m{#1}}
\newcolumntype{C}[1]{>{\centering\let\newline\\\arraybackslash\hspace{0pt}}m{#1}}
\newcolumntype{R}[1]{>{\raggedleft\let\newline\\\arraybackslash\hspace{0pt}}m{#1}}

\usepackage{tabularx}
\usepackage[table]{xcolor}
\usepackage{multirow}
\newcommand{\name}[0]{GeoLoco}

\newcommand\blfootnote[1]{% 
\begingroup 
\renewcommand\thefootnote{}\footnote{#1}% 
\addtocounter{footnote}{-1}% 
\endgroup 
}

\title{\LARGE \bf GeoLoco: Leveraging 3D Geometric Priors from Visual Foundation Model for Robust RGB-Only Humanoid Locomotion}
\author{Yufei Liu ~~  Xieyuanli Chen ~~ Hainan Pan ~~ Chenghao Shi ~~ Yanjie Chen \\ Kaihong Huang ~~ Zhiwen Zeng ~~ Huimin Lu*  %   <-this % stops a space
}

\begin{document}
\twocolumn[{%
\renewcommand\twocolumn[1][]{#1}%
\maketitle

\begin{figure}[H]
\hsize=\textwidth
  \centering
  \includegraphics[width=1.99\linewidth]{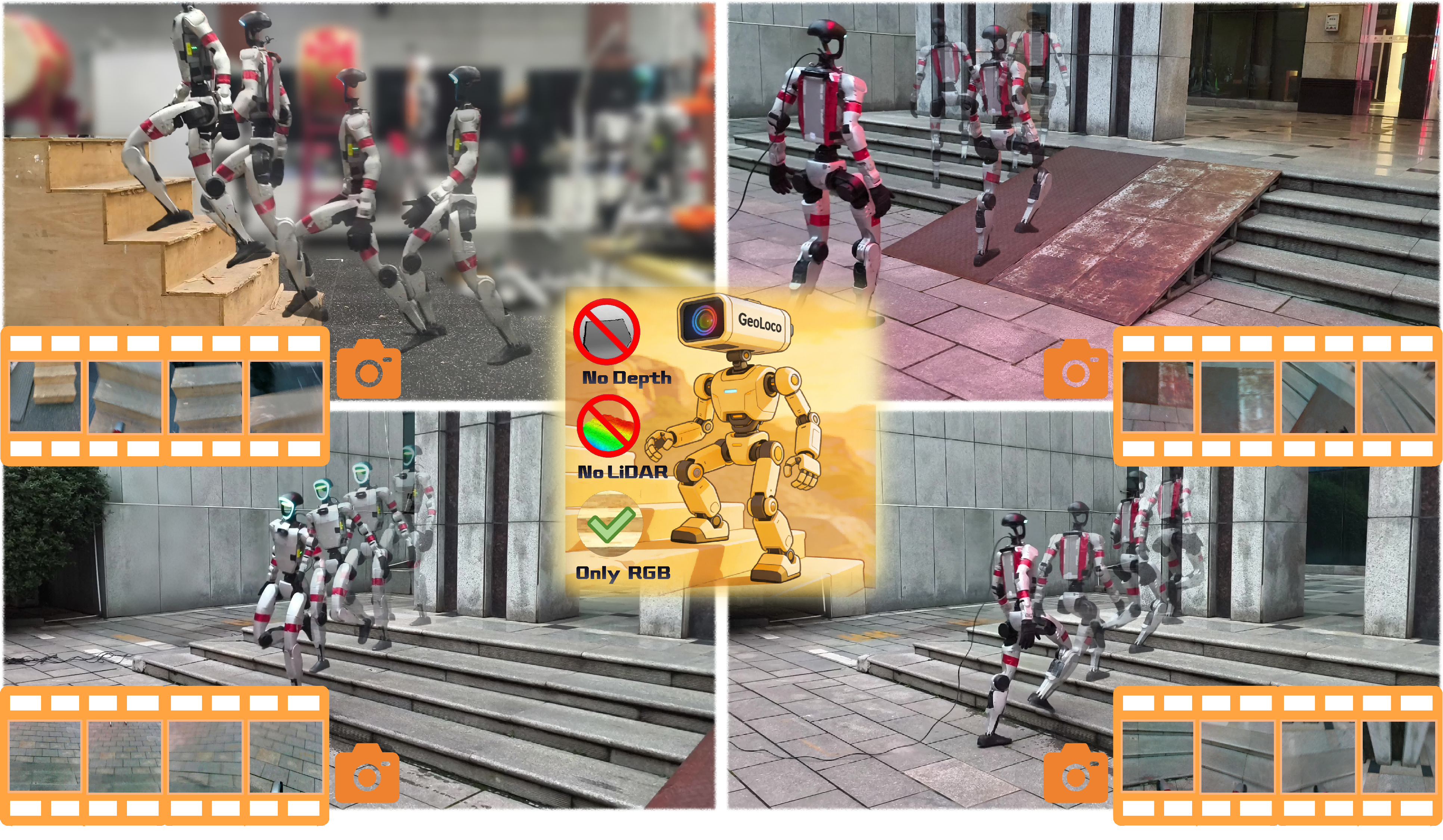}
  \caption{\textbf{Overview of \name{}.} Our framework enables robust humanoid locomotion on diverse terrains—including stairs, ramps, and uneven blocks—using \textbf{only a monocular RGB camera.} By conceptualizing 2D pixels as high-dimensional 3D geometric representations, \name{} eliminates the dependency on active depth sensors (LiDAR/Depth) while maintaining expert-level traversal performance.}
  \label{fig:overview}
\end{figure}
}]

\blfootnote{
All authors are with the College of Intelligence Science and Technology, National University of Defense Technology, China. 
% This work was supported in part by the National Science Foundation of China under Grant 62403478, Young Elite Scientists Sponsorship Program by CAST (No. 2023QNRC001).
}

\begin{abstract}
The prevailing paradigm of perceptive humanoid locomotion relies heavily on active depth sensors. However, this depth-centric approach fundamentally discards the rich semantic and dense appearance cues of the visual world, severing low-level control from the high-level reasoning essential for general embodied intelligence. While monocular RGB offers a ubiquitous, information-dense alternative, end-to-end reinforcement learning from raw 2D pixels suffers from extreme sample inefficiency and catastrophic sim-to-real collapse due to the inherent loss of geometric scale. To break this deadlock, we propose GeoLoco, a purely RGB-driven locomotion framework that conceptualizes monocular images as high-dimensional 3D latent representations by harnessing the powerful geometric priors of a frozen, scale-aware Visual Foundation Model (VFM). Rather than naive feature concatenation, we design a proprioceptive-query multi-head cross-attention mechanism that dynamically attends to task-critical topological features conditioned on the robot's real-time gait phase. Crucially, to prevent the policy from overfitting to superficial textures, we introduce a dual-head auxiliary learning scheme. This explicit regularization forces the high-dimensional latent space to strictly align with the physical terrain geometry, ensuring robust zero-shot sim-to-real transfer. Trained exclusively in simulation, GeoLoco achieves robust zero-shot transfer to the Unitree G1 humanoid and successfully negotiates challenging terrains.
\end{abstract}

% %%%%%%%%%%%%%%%%%%%%%%%%%%%%%%%%%%%%%%%%%%%%%%%%%%%%%%%%%%%%%%%%%%%%%%%%%%%%%%%%
% \keywords
% Humanoid locomotion, visual perception, Monocular RGB, reinforcement learning.
% \endkeywords
%%%%%%%%%%%%%%%%%%%%%%%%%%%%%%%%%%%%%%%%%%%%%%%%%%%%%%%%%%%%%%%%%%%%%%%%%%%%%%%%

\section{Introduction}
\label{sec:intro}

Humanoid locomotion represents a fundamental pillar of general embodied intelligence, requiring the seamless integration of high-level perception and low-level motor control. Recently, deep reinforcement learning (DRL) has significantly accelerated this field, enabling robots to exhibit remarkable agility and gait emergence within high-fidelity simulations \cite{Mittal2023isaaclab, viktor2021isaacgym}. However, bridging the gap to unstructured real-world environments remains a formidable challenge. While proprioception-only policies have demonstrated impressive disturbance rejection on flat or slightly uneven ground \cite{Aswin2023dreamwaq, Cui2024corl}, they remain fundamentally blind to upcoming geometric discontinuities such as stairs and discrete obstacles. Negotiating such terrains necessitates perceptive locomotion, where the control policy must actively reason about terrain geometry to proactively adjust foot placement and whole-body posture.

The prevailing paradigm for perceptive locomotion heavily relies on active depth sensors, such as LiDAR or RGB-D cameras, to explicitly reconstruct local elevation maps or height fields \cite{zhuang2024parkour, wang2025beamdojo}. Although effective for geometric stability, this depth-centric paradigm is increasingly becoming an information silo that is misaligned with the broader trend of integrated embodied intelligence. By stripping away dense appearance and semantic cues in favor of sparse geometric representations, these methods disconnect the low-level locomotion policy from the rich contextual information required for cross-task reasoning and Vision-Language-Action (VLA) alignment. This decoupling hinders the potential for locomotion policies to benefit from the scaling laws of modern Visual Foundation Models (VFMs). Consequently, transitioning to a purely RGB-driven paradigm inherently preserves this dense contextual richness, establishing a scalable prerequisite for integrating low-level locomotion into general-purpose VLA frameworks.

Despite the potential of the RGB modality, achieving robust geometry-aware locomotion from a single monocular camera remains a formidable challenge. Standard 2D pixel observations inherently lack metric scale and are highly susceptible to visual ambiguities and lighting variations, making end-to-end RL notoriously sample-inefficient and brittle during simulation-to-real transfer (sim-to-real). However, the emergence of large-scale VFMs offers a transformative perspective. In this work, we posit that monocular RGB input should be re-conceptualized not merely as a 2D pixel array, but as a latent representation of the 3D world, unlocked by the scaling laws and geometric priors embedded in modern VFMs.
By leveraging these pre-trained models, we can extract robust, metric-aware geometric cues that were fundamentally inaccessible to Convolutional Neural Network (CNN) baselines.

\begin{figure*}[t]
  \vspace{-0.3em}
  \centering
  \includegraphics[width=0.98\textwidth]{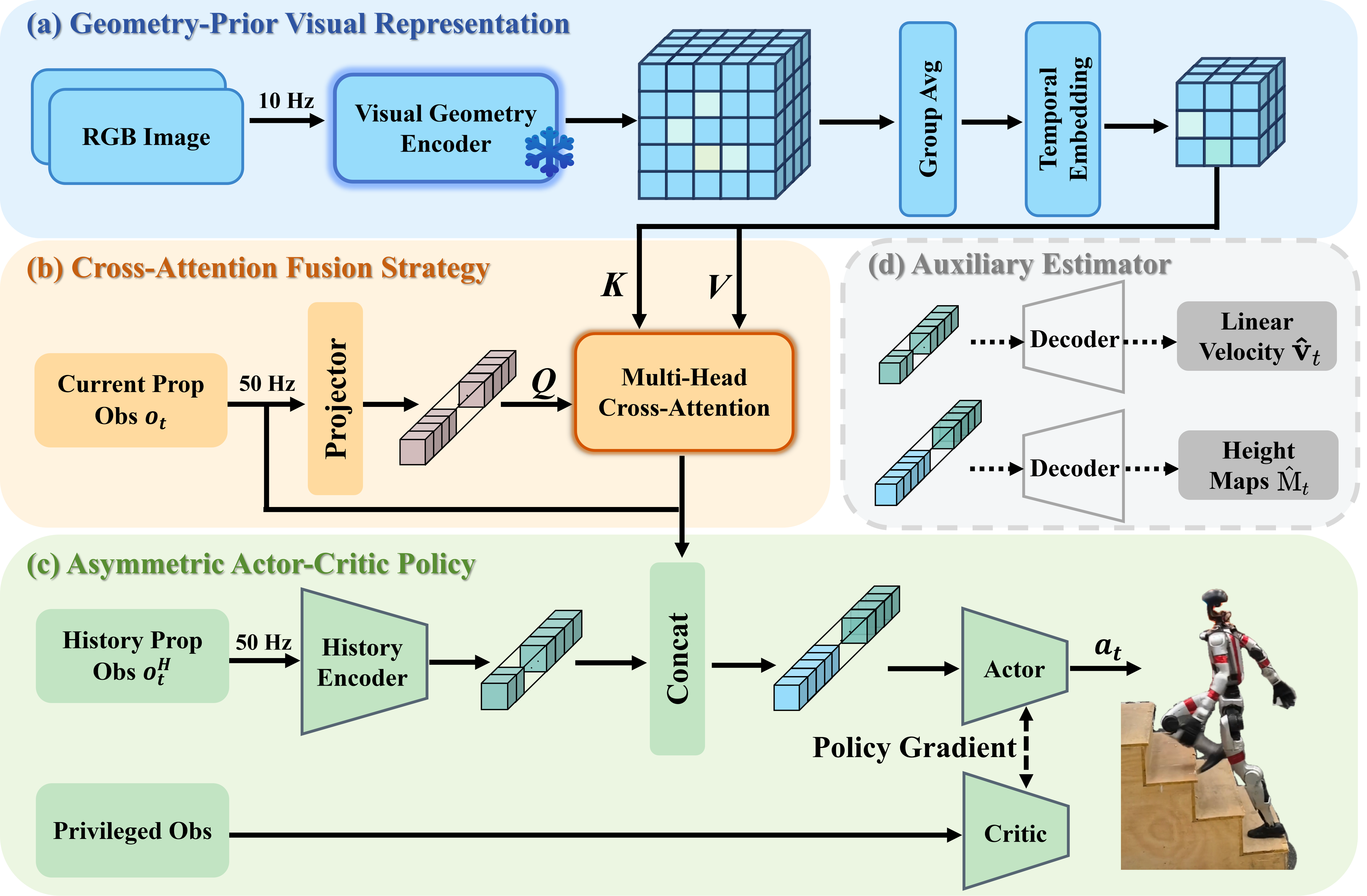}
  \caption{\textbf{The \name{} Architecture.} A frozen, scale-aware \textbf{Visual Geometry Encoder} extracts multi-scale 3D priors from asynchronous RGB streams (10\,Hz). These features are fused with high-frequency proprioceptive (50 \,Hz) via a \textbf{Multi-Head Cross-Attention mechanism.} To bridge the sim-to-real gap, a dual-head auxiliary decoder (middle right) regularizes the latent space by reconstructing local terrain topography and predicting system dynamics during training.}
  \label{fig:pipeline}
\end{figure*}

To address these challenges, we propose \name{}, a purely RGB-driven reinforcement learning framework for robust humanoid locomotion. Bypassing active depth sensors, \name{} employs a frozen, scale-aware visual geometry encoder to project monocular RGB observations into high-dimensional geometric features. 
As shown in \figref{fig:pipeline}, a lightweight cross-attention module fuses visual embeddings with proprioceptive observation. Guided by real-time kinematics, the policy dynamically focuses on critical terrain features like stair edges and riser heights.
Crucially, to mitigate overfitting to superficial visual textures, we introduce a dual-head auxiliary learning scheme. This explicit regularization physically grounds the high-dimensional latent space by enforcing strict alignment with local terrain map and robot states. Trained exclusively in simulation, \name{} achieves zero-shot deployment on the Unitree G1 humanoid, successfully traversing severe geometric discontinuities without any real-world fine-tuning.

The main contributions of this paper are as follows:
\begin{itemize}
    \item We propose a purely RGB-driven humanoid locomotion framework that bridges geometric precision with semantic richness by conceptualizing monocular RGB as a 3D latent representation, effectively eliminating the reliance on active depth sensors.
    \item We introduce a lightweight visual-proprioceptive fusion architecture utilizing cross-attention, which efficiently integrates high-dimensional visual priors with proprioceptive states to facilitate geometry-aware whole-body control.
    \item We design a dual-head auxiliary representation learning scheme that regularizes the latent space by reconstructing terrain topography and predicting system dynamics, ensuring robust zero-shot sim-to-real transfer under diverse environmental conditions.
\end{itemize}

%%%%%%%%%%%%%%%%%%%%%%%%%%%%%%%%%%%%%%%%%%%%%%%%%%%%%%%%%%%%%%%%%%%%%%%%%%%%%%%%
\section{Related Work}
\label{sec:related}

\subsection{Bottlenecks in Perceptive Locomotion}

While DRL has significantly enhanced humanoid agility~\cite{Mittal2023isaaclab}, proprioceptive-only policies~\cite{Aswin2023dreamwaq, Cui2024corl} remain fundamentally blind to geometric discontinuities. Consequently, \textit{perceptive locomotion} relying on active depth sensors has become indispensable~\cite{cheng2024icra, zhuang2024parkour}. However, this depth-centric paradigm creates a critical information silo. By discarding dense appearance and semantic cues, it severs low-level motor control from the high-level reasoning essential for general embodied intelligence. Transitioning to a purely RGB-driven modality is therefore inevitable for seamless Vision-Language-Action (VLA) alignment. 

\subsection{Visual Representations: From 2D Priors to 3D Priors}

Deriving metric geometry from monocular RGB remains a long-standing challenge~\cite{yang2024depthanythingv2}. Yet, extracting reliable, scale-aware geometry from monocular RGB remains a formidable bottleneck, explicitly necessitating a paradigm shift in visual representation. Recently, Visual Foundation Models (VFMs) have emerged as powerful feature extractors for robotics, driving advancements in humanoid whole-body control~\cite{xue2025leverb}, general-purpose manipulation~\cite{lin2025evo}, and Vision-Language Navigation (VLN)~\cite{zeng2025janusvln}. 

However, these applications predominantly focus on high-level semantics, task instructions, or quasi-static behaviors. For high-frequency perceptive locomotion over complex terrains, the critical bottlenecks are real-time inference latency and the acquisition of metric-scale geometry.

\subsection{Sim-to-Real Transfer for Visual Policy}

The primary obstacle to RGB-only control is the substantial sim-to-real gap in visual appearance. Recently, pure pixel-to-action RL has emerged as a promising paradigm for humanoid control. Pioneering works such as VIRAL~\cite{he2025viral} and related policy transfers~\cite{xue2025doorman} have demonstrated impressive visual sim-to-real capabilities at scale. Furthermore, end-to-end visual RL has even been shown to facilitate emergent active perception in simulated humanoids~\cite{luo2025pdc}. 

However, prevailing RGB-driven methods typically train visual encoders entirely from scratch, necessitating aggressive Domain Randomization (DR) that exacerbates sample inefficiency and overfitting to simulation artifacts. Conversely, \name{} integrates a frozen VFM prior with physically grounded auxiliary regularization. This formulation explicitly circumvents representation collapse, enabling highly sample-efficient training and robust zero-shot deployment in unstructured real-world environments.

%%%%%%%%%%%%%%%%%%%%%%%%%%%%%%%%%%%%%%%%%%%%%%%%%%%%%%%%%%%%%%%%%%%%%%%%%%%%%%%%
\section{Method}
\label{sec:main}

This section presents \name{}, a novel geometry-aware humanoid locomotion policy that relies exclusively on a monocular RGB camera and proprioception.

\subsection{Problem Formulation}
\label{sec:3.1}

We formulate the humanoid locomotion task as a Partially Observable Markov Decision Process (POMDP) solved via Proximal Policy Optimization (PPO)~\cite{schulman2017proximal}. At each control step $t$, the actor policy $\pi_{\theta}$ generates the joint position commands $\mathbf{a}_t \in \mathbb{R}^{d_a}$. The observation space consists of the instantaneous proprioceptive state $\mathbf{o}_t \in \mathbb{R}^{d_p}$, the target velocity command $\mathbf{c}_t \in \mathbb{R}^3$, and the monocular RGB image $\mathbf{I}_t \in \mathbb{R}^{112 \times 112 \times 3}$. To mitigate the inherent scale ambiguity of monocular vision and transient sensory occlusions, the policy is explicitly conditioned on a proprioceptive history of length $h$, defined as an ordered sequence $\mathbf{o}_t^H = (\mathbf{o}_{t-h+1}, \dots, \mathbf{o}_t)$. The critic has access to privileged simulator state $\mathbf{s}_t \in \mathbb{R}^{371}$ to provide richer value estimates during training. Include ground-truth base linear velocity and height, foot-end kinematics, 3D contact forces, and a dense $17\!\times\!11$ local elevation map $\mathbf{M}_t$.

To demonstrate the robustness of our RGB-driven visual representation, We employ a standard task-agnostic reward formulation prioritizing velocity tracking and joint smoothness, forcing the policy to rely strictly on 3D visual priors rather than reward engineering.

\subsection{Geometry-Prior Visual Representation from RGB}
\label{sec:geom_rgb}

To navigate complex terrains like stairs, a humanoid robot requires an awareness of the local 3D geometry that is invariant to lighting and texture. Instead of training a visual encoder from scratch, which often overfits to simulation-specific textures and fails in sim-to-real transfer, we leverage a pretrained VFM as a frozen 3D geometric prior. This paradigm shift allows the policy to operate on a high-dimensional representation that inherently encodes metric-relative depth and surface structures, effectively treating monocular RGB as a 3D latent surface rather than a flat 2D pixel array.

\subsubsection{Multi-Scale Tokenized Feature Extraction} 
Given a monocular RGB frame $\mathbf{I}_t \in \mathbb{R}^{112 \times 112 \times 3}$, the frozen VFM tokenizes it into $N = 64$ non-overlapping patches, arranged in an $8\!\times\!8$ spatial grid. We instantiate the VFM using Depth-Anything-V2 (ViT-S)~\cite{yang2024depthanythingv2} with embedding dimension $d_v\!=\!384$. Specifically, we leverage its metric-depth variant, which aligns perfectly with our objective of recovering absolute metric scale. Rather than using the scalar depth output, we extract the spatially-organized patch tokens from multiple intermediate transformer blocks $\ell$. 

Within the ViT-based architecture, shallow layers capture high-frequency geometric primitives, while deeper layers provide broader structural context~\cite{oquab2023dinov2,yang2024depthanythingv2}. To exploit this hierarchy, we extract tokens from layers $\ell \in \{4, 8, 12\}$ of the 12-block ViT-S encoder, yielding three feature maps, each with $N\!=\!64$ tokens of dimension $d_v\!=\!384$.

\begin{figure}[t]
  \centering
  \includegraphics[width=\linewidth]{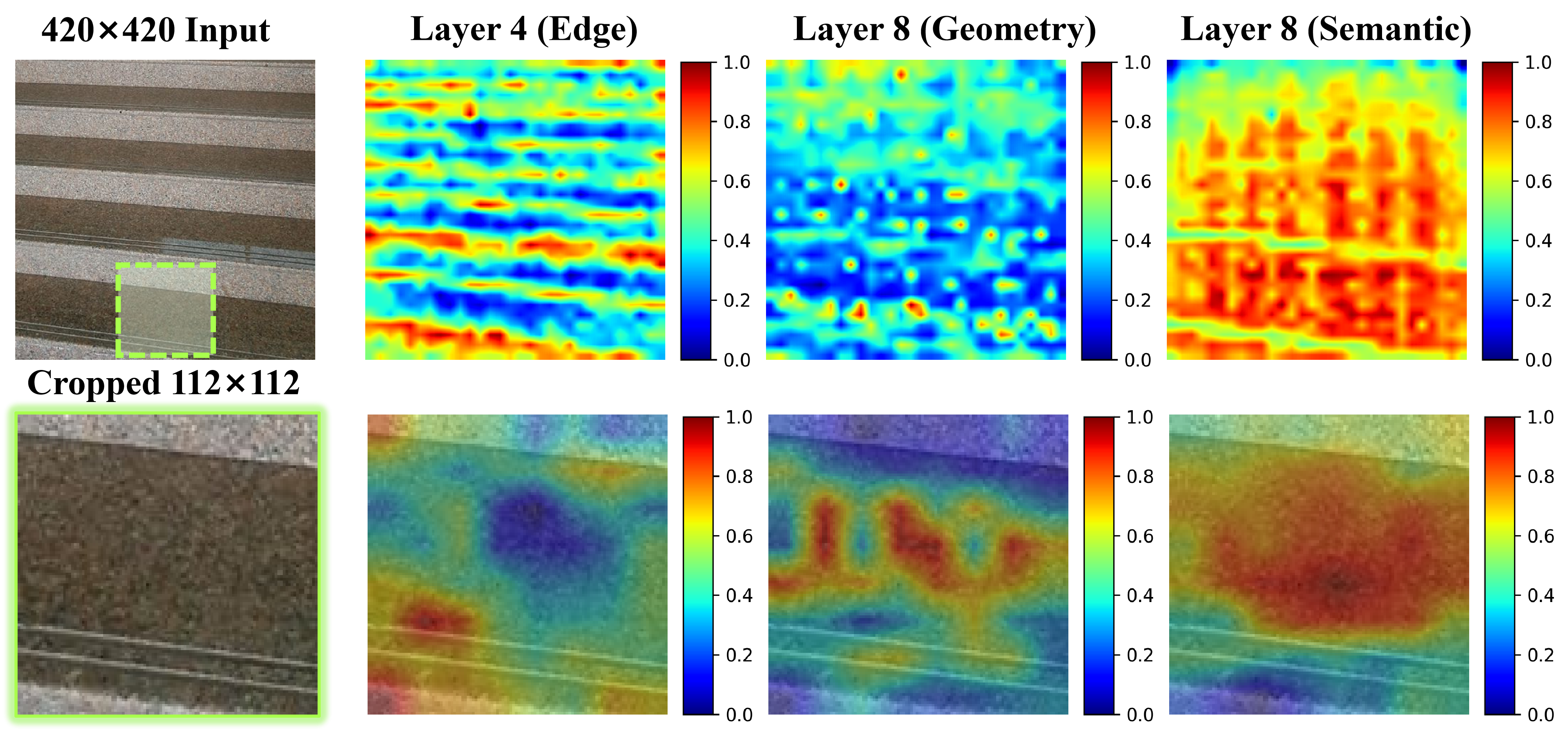}
  \caption{Multi-Scale geometric activations from the frozen VFM. We visualize patch token activations from intermediate transformer layers (4/8/12).}
  \label{fig:layer_activations}
\end{figure}

\subsubsection{Channel-Grouped Spatial Projection} 
Raw VFM embeddings occupy a high-dimensional semantic space unsuitable for high-frequency control. To adapt these representations into a compact spatial descriptor, we apply a parameter-free channel grouping operation independently to each layer. For each layer $\ell$, the $N$ patch tokens are first reshaped into a spatial feature map $\mathbf{F}^{(\ell)}_t \in \mathbb{R}^{d_v \times h \times w}$ where $h\!=\!w\!=\!8$. We then partition the $d_v\!=\!384$ channels into $G\!=\!32$ groups of 12 and average within each group:
\begin{equation}
\mathbf{Z}^{(\ell)}_t = \mathrm{GroupAvg}\!\left(\mathbf{F}^{(\ell)}_t\right) \in \mathbb{R}^{G \times h \times w}.
\label{eq:group_avg}
\end{equation}
The three layers are then concatenated along the channel dimension to form the multi-scale geometric descriptor:
\begin{equation}
\mathbf{T}_t = \mathrm{Concat}\!\left(\mathbf{Z}^{(4)}_t, \mathbf{Z}^{(8)}_t, \mathbf{Z}^{(12)}_t\right),
\label{eq:multilevel_tokens}
\end{equation}
yielding $\mathbf{T}_t \in \mathbb{R}^{96 \times 8 \times 8}$, a compact spatial feature map where each of the $N\!=\!64$ spatial locations carries a $d\!=\!3G\!=\!96$ dimensional descriptor. Compared to the raw concatenated ViT output, this represents an aggressive $12\times$ per-token compression while strictly preserving the $8\!\times\!8$ spatial layout critical for terrain geometry. Crucially, this operation is entirely parameter-free and operates within the frozen encoder, introducing no additional learnable weights.

\subsubsection{Temporal Alignment and Asynchronous Inference} 
A pivotal challenge in RGB-driven locomotion is the computational latency of large-scale VFMs. While the proprioceptive state $\mathbf{o}_t$ is sampled at $50\,\mathrm{Hz}$ for reactive balancing, the VFM inference is decoupled and runs asynchronously at $10\,\mathrm{Hz}$. Let $\Delta t_{\mathrm{vis}}$ denote the visual update interval in terms of control steps. For any high-frequency control step $t$, the policy consumes the most recently cached visual descriptor $\mathbf{T}_{\tau}$, where the visual timestamp $\tau$ is updated via a zero-order hold mechanism:
\begin{equation}
\tau = \left\lfloor \frac{t}{\Delta t_{\mathrm{vis}}} \right\rfloor \Delta t_{\mathrm{vis}}.
\label{eq:async_tau}
\end{equation}
This asynchronous design ensures that the humanoid's reactive control loop remains unconstrained by the visual processing latency, while consistently benefiting from periodic 3D geometric corrections.

\begin{figure}[t]
  \centering
  \includegraphics[width=0.95\linewidth]{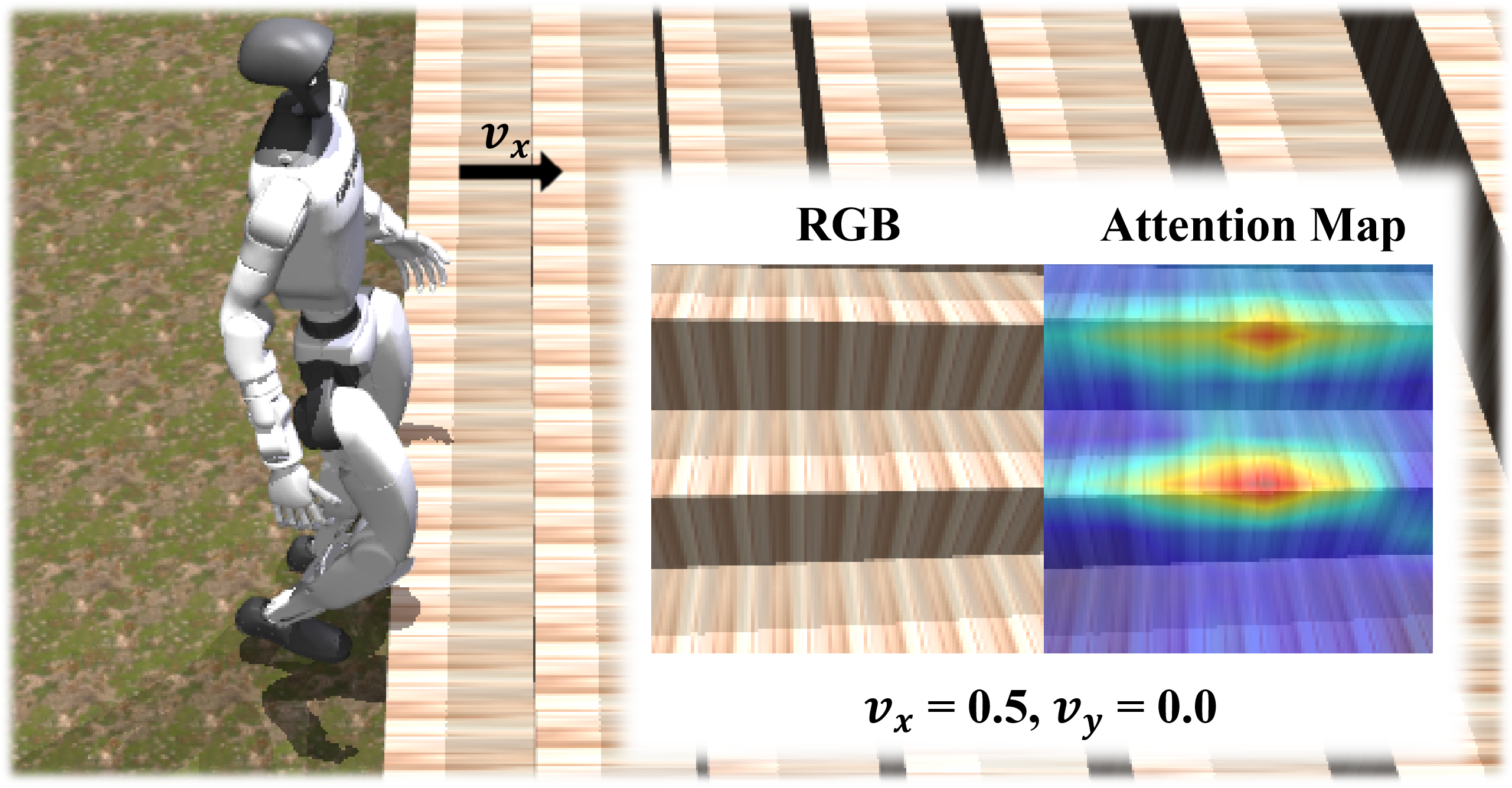}
  \vspace{-0.4em}
  \caption{Visualization of attention heatmap produced by the proprioceptive-query cross-attention module for a representative stair scene.}
  \vspace{-0.8em}
  \label{fig:terrain}
\end{figure}
% \subsection{Cross-Attention Fusion Strategy with Geometry-Consistent Aux Loss}

\subsection{Multi-Head Cross-Attention Fusion}
\label{sec:fusion}

To bridge the gap between high-dimensional geometric representations and reactive motor control, we propose a Multi-Head Cross-Attention mechanism. Unlike conventional methods utilizing static convolution or naive concatenation, our approach treats humanoid locomotion as an active perception process, where the robot's instantaneous physical state dynamically modulates its visual focus.

\subsubsection{Spatio-Temporal Token Formulation}
Instead of reasoning over a single frame, we aggregate geometry-prior descriptors from a short visual window of the most recent two asynchronous updates. The spatial feature map $\mathbf{T}_{\tau}$ from each frame $f$ is flattened along its spatial dimensions to yield $N\!=\!64$ sequence tokens, each of dimension $d\!=\!96$. To enable the attention mechanism to distinguish tokens across time, we inject a learnable temporal embedding $\mathbf{e}_{\mathrm{frame}}(f) \in \mathbb{R}^{d}$ into each token:
\begin{equation}
    \tilde{\mathbf{t}}_{i,f} = \mathbf{t}_{i,f} + \mathbf{e}_{\mathrm{frame}}(f), \quad i \in \{1,\dots,N\}, \; f \in \{0,1\}.
    \label{eq:token_embedding}
\end{equation}
These tokens are then concatenated to form a single spatio-temporal sequence $\tilde{\mathbf{T}}_{\tau} \in \mathbb{R}^{M \times d}$, where $M = 128$. This formulation empowers the subsequent attention mechanism to perform patch-level temporal alignment and deduce the relative motion of terrain features.

\subsubsection{Proprioceptive-Query Mechanism}
Our fusion formulation posits that the robot's instantaneous kinematic and dynamic state must actively modulate spatial attention. Consequently, the instantaneous proprioceptive observation $\mathbf{o}_t \in \mathbb{R}^{d_o}$ ($d_o\!=\!96$) is explicitly projected as the attention query $\mathbf{q}_t$, whereas the compressed proprioceptive history $\mathbf{h}_t$ is decoupled to govern high-level policy execution and velocity estimation.

We project the current state into a query $\mathbf{q}_t$, and the enriched visual tokens into keys $\mathbf{K}_{\tau}$ and values $\mathbf{V}_{\tau}$:
\begin{equation}
\begin{split}
    \mathbf{q}_t &= \mathbf{W}_\mathrm{Q} \mathbf{o}_t \in \mathbb{R}^{1 \times d_a}, \\
    \mathbf{K}_{\tau} &= \tilde{\mathbf{T}}_{\tau} \mathbf{W}_\mathrm{K} \in \mathbb{R}^{M \times d_a}, \quad \mathbf{V}_{\tau} = \tilde{\mathbf{T}}_{\tau} \mathbf{W}_\mathrm{V} \in \mathbb{R}^{M \times d_a},
\end{split}
\label{eq:qkv_new}
\end{equation}
where $d_a\!=\!64$ is the attention embedding dimension, with $\mathbf{W}_\mathrm{Q} \in \mathbb{R}^{d_o \times d_a}$ and $\mathbf{W}_\mathrm{K}, \mathbf{W}_\mathrm{V} \in \mathbb{R}^{d \times d_a}$. We employ multi-head attention (MHA) with $8$ heads to allow parallel focus on multiple task-critical spatial regions. For the $j$-th attention head, the localized attention weights $\boldsymbol{\alpha}^{(j)}_t$ are computed as:
\begin{equation}
    \boldsymbol{\alpha}^{(j)}_t = \mathrm{softmax}\!\left(\frac{\mathbf{q}^{(j)}_t (\mathbf{K}^{(j)}_{\tau})^\top}{\sqrt{d_{\mathrm{head}}}}\right) \in \mathbb{R}^{1 \times M}.
    \label{eq:cross_attn_final}
\end{equation}
The outputs from all heads are concatenated and projected through a final linear layer with ELU activation, producing the fused visual context $\mathbf{c}^{\mathrm{atten}}_t \in \mathbb{R}^{d_z}$.

\subsubsection{Physical Interpretation of the Attention Score}
Our formulation inherently grounds the attention weights $\boldsymbol{\alpha}_t$ into three physical properties: semantic matching to extract traversable geometric primitives, spatial preference to bias visual focus based on postural inclination, and temporal preference to weigh historical versus instantaneous frames conditioned on locomotion velocity.

The final policy input is formed by concatenating the instantaneous state, the temporal proprioceptive history, and the state-dependent visual context: $\mathbf{u}_t^{\mathrm{in}} = [\mathbf{o}_t;\; \mathbf{h}_t;\; \mathbf{c}^{\mathrm{atten}}_t] \in \mathbb{R}^{288}$, ensuring a clean decoupling between reactive balance and geometry-aware path planning.

\subsection{Geometry-Aware Auxiliary Estimator}
\label{sec:aux_loss}

To ensure that the fused latent space captures metric-accurate geometry rather than mere visual textures, we introduce an auxiliary representation learning scheme. During training, two lightweight decoding heads provide direct geometric supervision to the proprioceptive and visual pathways:

\begin{itemize}
    \item \textbf{Velocity Estimation Head:} Estimates the robot's base linear velocity $\hat{\mathbf{v}}_t = g_{\mathrm{vel}}(\mathbf{h}_t) \in \mathbb{R}^{3}$ from the history latent alone. This provides an explicit supervisory gradient to the history encoder, complementary to the implicit RL policy gradient.
    \item \textbf{Terrain Reconstruction Head:} Reconstructs a front-facing subset of the local height map $\hat{\mathbf{M}}_t = g_{\mathrm{recon}}(\mathbf{u}_t^{\mathrm{in}}) \in \mathbb{R}^{99}$ from the full policy input. Only the forward half ($x \ge 0$) of the grid is predicted, focusing supervision on the terrain the robot is about to traverse.
\end{itemize}

The corresponding ground-truth signals $\mathbf{v}_t$ and $\mathbf{M}_t$ are obtained directly from the simulator. We define the total objective function as a combination of the PPO surrogate loss and the auxiliary regularizers:
\begin{equation}
\mathcal{L} = \mathcal{L}_{\mathrm{PPO}} + \lambda_{\mathrm{vel}} \mathrm{MSE}(\hat{\mathbf{v}}_t, \mathbf{v}_t) + \lambda_{\mathrm{recon}} \mathrm{MSE}(\hat{\mathbf{M}}_t, \mathbf{M}_t),
\end{equation}
where $\lambda_{\mathrm{vel}}$ and $\lambda_{\mathrm{recon}}$ are weighting coefficients. At deployment, the auxiliary heads are discarded, introducing zero computational overhead.

\begin{table}[t]
\caption{Visual Domain Randomization Parameters}
\label{tab:visual_dr}
\begin{center}
\begin{tabular}{lcc}
\toprule
\textbf{Parameter} & \textbf{Randomization Range} & \textbf{Unit} \\
\midrule
Dome Light Intensity & $[300, 2000]$ & lux \\
Directional Light Intensity & $[1000, 6000]$ & lux \\
Light Color Variation & $\pm 30$ & \%  \\
Lighting Angles & $[0, 2\pi]$ & rad \\
Terrain Material & Random Texture Set & -- \\
Camera Exposure/Contrast & $[0.8, 1.2]$ & factor \\
Camera Position (x) & $[-10, 10]$ & mm \\
Camera Position (y) & $[-10, 10]$ & mm \\
Camera Position (z) & $[-10, 10]$ & mm \\
Motion Blur Probability & $15$ & \% \\
Random Latency & $[0, 100]$ & ms \\
\bottomrule
\end{tabular}
\end{center}
\end{table}

\begin{figure}[t]
  \centering
  \includegraphics[width=0.9\linewidth]{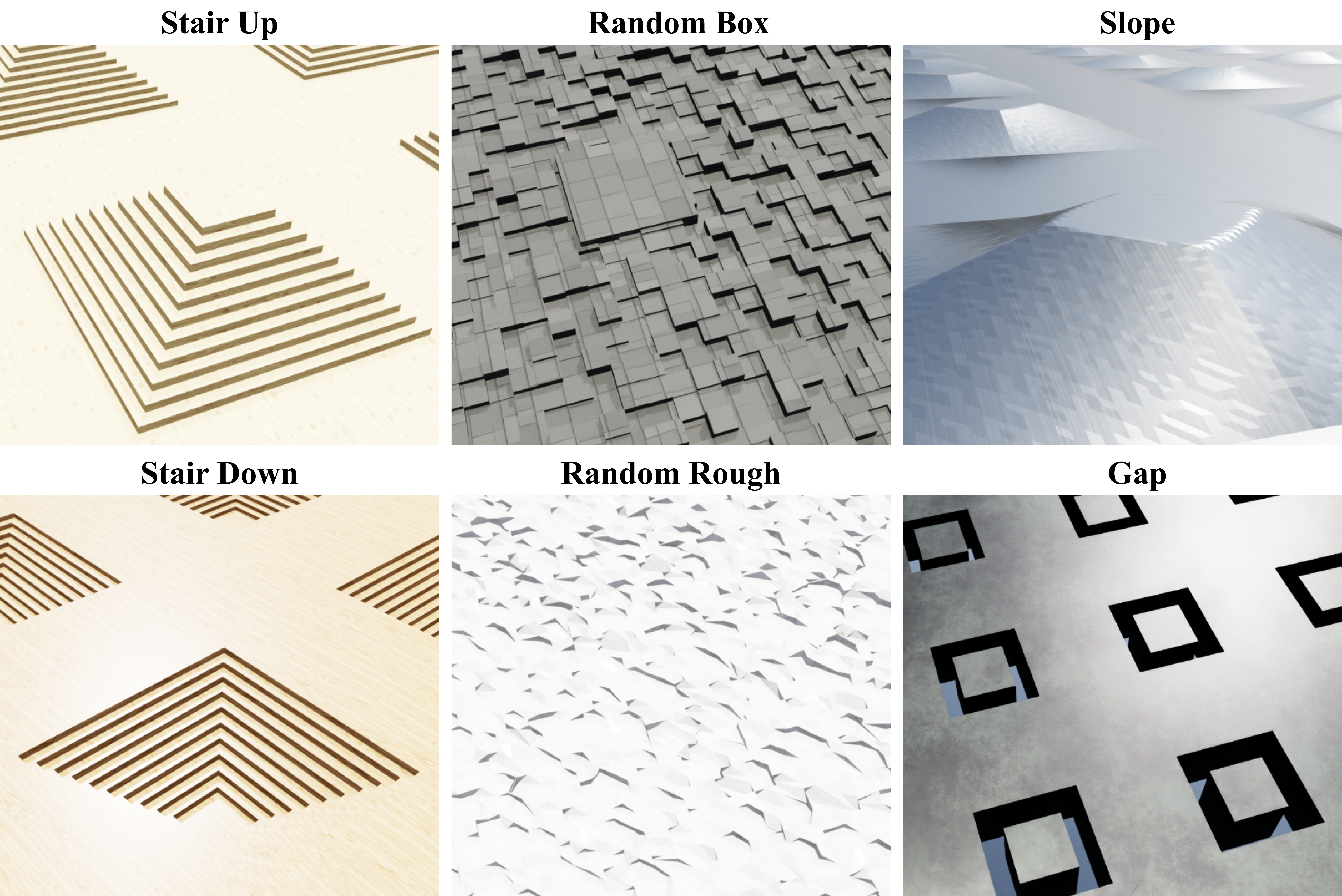}
  \vspace{-0.4em}
\caption{Training environments featuring diverse geometries and randomized textures. This prevents overfitting to visual appearances, ensuring the policy learns robust 3D priors for sim-to-real transfer.}
  \vspace{-0.8em}
  \label{fig:terrain}
\end{figure}

\begin{table*}[!t]
  \centering
  \caption{Benchmarked Comparison in Simulation. We compare \name{} against blind, depth-based, and other RGB-based baselines across various terrain difficulties. Best RGB-based result in bold.}
  \label{tab:success-rate}
  \renewcommand{\arraystretch}{1}
  \setlength{\tabcolsep}{3pt}
  \small
  \begin{tabularx}{\textwidth}{l l *{8}{>{\centering\arraybackslash}X}}
    \midrule
    \multirow{2}{*}{\textbf{Method}} & \multirow{2}{*}{\textbf{Modality}}
    & \multicolumn{2}{c}{Stairs-Up $\uparrow$}
    & \multicolumn{2}{c}{Stairs Down}
    & \multicolumn{2}{c}{Slope}
    & \multicolumn{2}{c}{Gap} \\
    \cline{3-4}\cline{5-6}\cline{7-8}\cline{9-10}
    & & \mbox{$R_{\mathrm{succ}}(\%, \uparrow)$} & \mbox{$R_{\mathrm{trav}}(\%, \uparrow)$}
        & \mbox{$R_{\mathrm{succ}}(\%, \uparrow)$} & \mbox{$R_{\mathrm{trav}}(\%, \uparrow)$}
        & \mbox{$R_{\mathrm{succ}}(\%, \uparrow)$} & \mbox{$R_{\mathrm{trav}}(\%, \uparrow)$}
        & \mbox{$R_{\mathrm{succ}}(\%, \uparrow)$} & \mbox{$R_{\mathrm{trav}}(\%, \uparrow)$} \\
    \midrule
    \rowcolor{black!8}\multicolumn{10}{l}{\textbf{\textit{Medium Terrain Difficulty}}} \\
    DreamWaq~\cite{Aswin2023dreamwaq} & Blind & 41.28 & 44.10 & 61.73 & 64.95 & 79.62 & 82.88 & 3.84 & 4.92 \\
    PIE~\cite{luo2024pie} & Depth & 90.63 & 92.10 & 93.22 & 94.58 & 98.91 & 99.42 & 92.44 & 93.76 \\
    MoRE~\cite{wang2025more} & Depth & 81.94 & 83.21 & 84.10 & 85.67 & 96.72 & 97.88 & 82.35 & 83.90 \\
    PLANC~\cite{dai2026planc} & Height map & 91.37 & 92.76 & 94.05 & 95.44 & 99.36 & 99.72 & 93.18 & 94.92 \\
    \midrule
    CNN & RGB & 59.42 & 60.88 & 70.35 & 71.90 & 84.72 & 85.96 & 29.58 & 30.74 \\
    GaussGym~\cite{Escontrela2025GaussGym} & RGB & 72.18 & 73.42 & 75.26 & 76.61 & 84.92 & 86.05 & 57.84 & 58.97 \\
    \textbf{\name{}} & RGB & \textbf{82.76} & \textbf{84.09} & \textbf{85.33} & \textbf{86.62} & \textbf{97.10} & \textbf{98.25} & \textbf{83.05} & \textbf{84.66} \\
    \midrule
    \rowcolor{black!8}\multicolumn{10}{l}{\textbf{\textit{Hard Terrain Difficulty}}} \\
    DreamWaq~\cite{Aswin2023dreamwaq} & Blind & 11.24 & 13.06 & 29.87 & 32.41 & 58.92 & 62.17 & 0.00 & 0.00 \\
    PIE~\cite{luo2024pie} & Depth & 84.78 & 86.25 & 88.91 & 90.34 & 98.36 & 99.10 & 61.52 & 62.89 \\
    MoRE~\cite{wang2025more} & Depth & 70.41 & 71.88 & 72.36 & 73.91 & 85.77 & 87.09 & 54.86 & 56.20 \\
    PLANC~\cite{dai2026planc} & Height map & 86.12 & 87.54 & 90.23 & 91.65 & 99.08 & 99.62 & 64.37 & 65.88 \\
    \midrule
    CNN & RGB & 28.11 & 30.19 & 60.38 & 61.90 & 69.55 & 70.88 & 17.86 & 19.40 \\
    GaussGym~\cite{Escontrela2025GaussGym} & RGB & 49.42 & 51.88 & 70.15 & 71.62 & 74.38 & 75.91 & 24.86 & 25.72 \\
    \textbf{\name{}} & RGB & \textbf{66.27} & \textbf{67.91} & \textbf{64.73} & \textbf{66.15} & \textbf{90.41} & \textbf{91.68} & \textbf{49.62} & \textbf{51.08} \\
    \hline
  \end{tabularx}
\end{table*}

\subsection{Sim-to-Real Training for RGB Policy}
\label{sec:sim_to_real}

A primary hurdle in transitioning from simulation to real-world deployment for RGB-based policies is the significant discrepancy in visual appearance and hardware dynamics. To ensure the zero-shot transferability of \name{}, we employ a comprehensive domain randomization strategy encompassing both physical properties and visual modalities.

% \subsubsection{Visual Domain Randomization}
To force the policy to prioritize the underlying 3D geometric structures over superficial 2D textures, we apply extensive randomization to the visual environment in IsaacLab~\cite{Mittal2023isaaclab}. By varying lighting intensities, color temperatures, and shadow directions shown in \tabref{tab:visual_dr}, we encourage the policy to extract lighting-invariant features. Furthermore, we randomly switch terrain materials during training, which prevents the agent from overfitting to specific simulation artifacts and ensures it relies on the metric-relative depth cues provided by the VFM.
Critically, to account for the asynchronous 10\,Hz visual update frequency against the 50\,Hz control loop, we inject random latency $[0, 100]$~ms into the visual feature stream.

\begin{figure*}[t]
  \centering
  \includegraphics[width=0.95\linewidth]{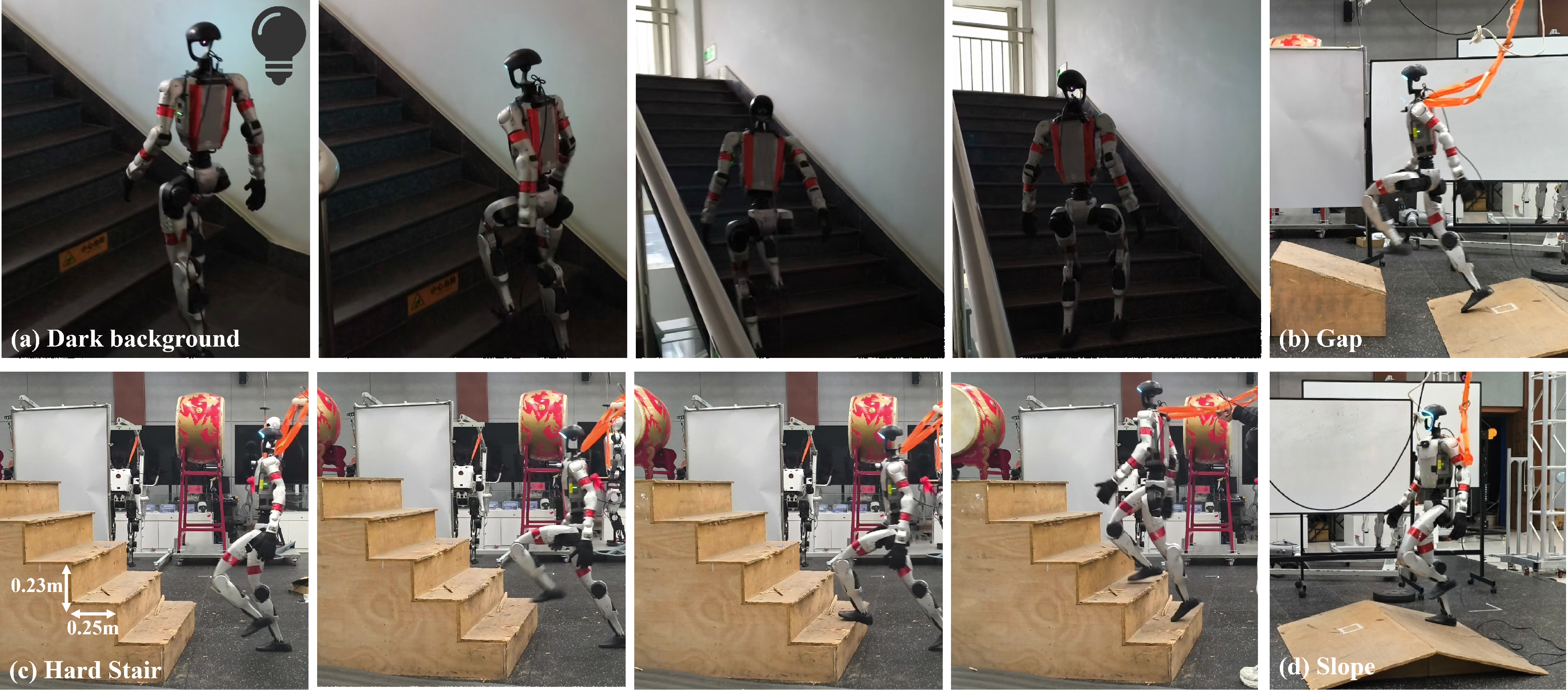}
  \vspace{-0.4em}
  \caption{\textbf{Real-world zero-shot deployment on Unitree G1.} The policy successfully traverses challenging geometric discontinuities under varied conditions, including (a) dark background stairs, (b) a gap obstacle, (c) high-riser stairs, and (d) a slope.}
  \vspace{-0.8em}
  \label{fig:real_world}
\end{figure*}

\begin{table*}[t]
  \caption{Quantitative ablation results in simulation. We report the maximum Mean Difficulty Level reached through curriculum learning and the corresponding Success Rate $R_{\mathrm{succ}}\%$ on the final curriculum stage. All results are averaged over 5 random seeds.}
  \label{tab:ablation_sim}
  \centering
  \footnotesize
  \setlength{\tabcolsep}{12pt}
  \renewcommand{\arraystretch}{1}
  \begin{tabular}{llcccc}
    \toprule
    Category & Variant & Modality & Fusion Strategy & Difficulty $\uparrow$ & \mbox{$R_{\mathrm{succ}}(\%, \uparrow)$} \\
    \midrule
    \textbf{Full Method} & \textbf{\name{} (Ours)} & RGB & \textbf{Cross-Attention} & \textbf{0.87} & \textbf{86.4} \\
    \midrule
    \multirow{4}{*}{(A) Representation} 
    & Scratch CNN (End-to-End) & RGB & Cross-Attention & 0.58 & 60.4 \\
    & Semantic VFM (DINOv2) & RGB & Cross-Attention & 0.61 & 66.1\\
    & CLS Tokens & RGB & Cross-Attention & 0.55 & 58.7 \\
    & Single-Layer Patch Token & RGB & Cross-Attention & 0.75 & 74.8 \\
    
    \midrule
    \multirow{4}{*}{(B) Modality Fusion} 
    & Naive MLP Concatenation & RGB & Concatenation & 0.76 & 71.2 \\
    & w/o Proprioceptive History ($h_t$) & RGB & Cross-Attention & 0.64 & 67.2 \\
    & Single-frame Vision (10\,Hz) & RGB & Concatenation & 0.77 & 79.5 \\
    & Temporal Vision (10\,Hz) & RGB & Conv2d & 0.79 & 83.5 \\
    \midrule
    \multirow{3}{*}{(C) Regularization} 
    & w/o Terrain Reconstruction & RGB & Cross-Attention & 0.71 & 74.2 \\
    & w/o Velocity Estimation  & RGB & Cross-Attention & 0.81 & 81.4 \\
    & w/o Visual Domain Randomization & RGB & Cross-Attention & 0.80 & 84.1 \\
    
    \midrule
    \multirow{3}{*}{(D) Baselines} 
    & Proprioceptive Only (Blind) & -- & None & 0.31 & 20.1 \\
    & Depth-based baseline \cite{wang2025more} & Depth & Concatenation & 0.94 & 93.2 \\
    & Height maps \cite{zhuang2024parkour} & LiDAR & Concatenation & 0.95 & 94.5 \\
    
    \bottomrule
  \end{tabular}
\end{table*}

\section{Experiments}
\label{sec:exp}

\subsection{Experimental Setup}
\label{sec:exp:setup}

We compare \name{} with three categories of baselines: {Proprioception-only DreamWaq~\cite{Aswin2023dreamwaq})}: Represents the lower bound without terrain awareness; {Active Geometric Sensors (PIE~\cite{luo2024pie}, MORE~\cite{wang2025more}, PLANC~\cite{dai2026planc})}: Use depth maps or height fields, representing the current perceptive locomotion standard; and {RGB-based End-to-End (CNN, GaussGym~\cite{Escontrela2025GaussGym})}. The training and simulation environments are implemented in IsaacLab~\cite{Mittal2023isaaclab}. Following standard perceptive locomotion benchmarks~\cite{wang2025beamdojo}, we quantitatively evaluate the policies using two primary metrics:
\begin{itemize}
    \item \textbf{Success Rate ($R_{\mathrm{succ}}$):} The percentage of episodes where the humanoid successfully traverses the entire predefined length of the target terrain without triggering any termination conditions.
    \item \textbf{Traversal Rate ($R_{\mathrm{trav}}$):} The average ratio of the forward distance traveled before an episode terminates to the total length of the terrain. This metric provides a soft penalty for partial successes.
\end{itemize}

\textbf{Evaluation Protocol:} During evaluation, we leverage massively parallel simulation, deploying $1,024$ independent environments for each terrain setting. Each environment operates continuously and resets automatically upon episode termination. This process iterates until 5 complete trials are collected per environment. Our curriculum scales difficulty from $0.0$ to $1.0$ (e.g., max stair height $0.25\,\mathrm{m}$, max gap $0.3\,\mathrm{m}$). To systematically probe the robustness boundaries of different modalities, we evaluate on two distinct tiers: \textit{Medium} (level $0.4$) and \textit{Hard} (level $0.8$).

\textbf{Network Architecture and Implementation Details:} 
For the visual perception pipeline, we adopt the small variant of the Depth-Anything-V2 (Metric Depth)~\cite{yang2024depthanythingv2} model as our frozen VFM backbone. The model utilizes a Vision Transformer (ViT-S) encoder~\cite{dosovitskiy2021vit} with approximately $24.8\,\mathrm{M}$ parameters. For the proprioceptive history, we set the window length to $h = 5$.

\subsection{Simulation Results}
\label{sec:exp:sim}

We report the quantitative evaluation across varying terrain difficulties in~\tabref{tab:success-rate}. The results reveal several critical insights regarding sensory modalities and visual representation learning in perceptive locomotion. First, the proprioceptive-only baseline~\cite{Aswin2023dreamwaq} completely fails on severe geometric discontinuities, yielding a $0.00\%$ success rate $R_{\mathrm{succ}}$ on Hard Gaps. This confirms the absolute necessity of exteroception for negotiating non-continuous terrains.

Second, standard RGB-driven policies exhibit severe vulnerability to environmental complexity. While end-to-end CNN and GaussGym~\cite{Escontrela2025GaussGym} demonstrate moderate success on Medium terrains, their performance degrades sharply under Hard conditions. In contrast, \name{} substantially mitigates this degradation, maintaining a robust $R_{\mathrm{succ}}$ of $66.27\%$ on Hard Stairs-Up and $49.62\%$ on Hard Gaps. This substantial margin validates that conceptualizing monocular RGB as a 3D latent representation explicitly recovers the metric-level geometric cues inherently lacking in raw 2D pixel observations.

Finally, we benchmark \name{} against depth and heightmap-based methods, which utilize explicit geometric sensors. Remarkably, our purely RGB-driven framework achieves performance competitive with the depth-based MoRE baseline~\cite{wang2025more}. On Medium Stairs-Up, \name{} ($82.76\%$) even marginally exceeds MoRE ($81.94\%$). A predictable performance gap remains between our method and explicit heightmap algorithms on Hard terrains due to the physical limitations of monocular scale ambiguity.

\subsection{Real-World Deployment on Unitree G1}

\begin{table}[t]
\caption{Quantitative Real-World Performance on Unitree G1. Each method was evaluated $10$ trials per terrain scenario.  Best result in bold.}
\label{tab:real_world_quant}
\centering
\resizebox{\columnwidth}{!}{%
\begin{tabular}{l|cc|cc}
\toprule
\multirow{2}{*}{Method} & \multicolumn{2}{c|}{Stairs ($0.23\,\mathrm{m}$ riser)} & \multicolumn{2}{c}{Gap ($0.25\,\mathrm{m}$ width)} \\
 & $R_{\mathrm{succ}}(\%, \uparrow)$ & Time ($\downarrow$) & $R_{\mathrm{succ}}(\%, \uparrow)$ & Time ($\downarrow$) \\
\midrule
Proprioceptive (Blind)  & $30\%$   & $10.1$       & $0\%$  & $6.5$   \\
CNN-based   & $40\%$  & $8.2$   & $40\%$  & $5.8$   \\
\rowcolor{gray!15} \textbf{\name{} (Ours)} & \textbf{80\%} & \textbf{5.4} & \textbf{70\%} & \textbf{4.2} \\
\bottomrule
\end{tabular}%
}
\end{table}
We deploy the trained \name{} policy zero-shot on the 29-DoF Unitree G1 humanoid. The control loop operates at $50\,\mathrm{Hz}$, mirroring the simulation environment. Exteroception relies solely on the RGB stream of an onboard Intel RealSense D435i camera, with policy inference executed on a desktop RTX 4090 GPU.

\textbf{Quantitative Evaluation.}
To rigorously assess real-world reliability, we benchmark \name{} against both blind and end-to-end CNN baselines across 10 consecutive trials per terrain, as summarized in ~\tabref{tab:real_world_quant}. \name{} achieves success rates of $80\%$ and $70\%$ on $0.23\,\mathrm{m}$ stairs and $0.25\,\mathrm{m}$ gaps, respectively, substantially outperforming the baselines. Furthermore, the VFM-derived geometric priors enable proactive traversals, consistently reducing the average execution time to $5.4\,\mathrm{s}$ on stairs and $4.2\,\mathrm{s}$ on gaps.

\textbf{Qualitative Dynamic Behaviors.}
Qualitatively, the physical deployment validates several robust dynamic capabilities. On severe geometric discontinuities like steep stairs and wide gaps shown in~\figref{fig:real_world}b, c, the robot refrains from naive velocity tracking, exhibiting instead anticipatory leg-lifting and adaptive foot placement. Furthermore, \name{} successfully negotiates stairs under low-light conditions (\figref{fig:real_world}a), confirming that the extracted 3D latent representations inherently resist superficial illumination shifts. Finally, when traversing continuous slopes (\figref{fig:real_world}d), the policy dynamically regulates the torso pitch and center of mass, achieving geometry-aware whole-body coordination without explicit terrain-specific reward engineering.

\subsection{Ablation Analysis}
\label{sec:exp:ablation}

Table \ref{tab:ablation_sim} presents extensive ablations to isolate the contributions of three core components: 

\subsubsection{Efficacy of 3D Geometric Priors}
% A foundational premise of \name{} is the necessity of conceptualizing monocular RGB as a 3D latent representation rather than a flat 2D pixel array. 
As reported in \tabref{tab:ablation_sim}(A), replacing the frozen geometric encoder with an end-to-end trained  CNN precipitates a severe performance degradation, reducing the success rate $R_{\mathrm{succ}}$ from $86.4\%$ to $60.4\%$. Furthermore, employing a state-of-the-art semantic foundation model DINOv2~\cite{oquab2023dinov2} yields marginal benefits, achieving an $R_{\mathrm{succ}}$ of only $66.1\%$. This substantial performance gap substantiates that while semantic models excel at object-level recognition, they fundamentally lack the precise metric-scale spatial grounding required for negotiating discrete geometric discontinuities.

\subsubsection{Impact of Multi-Head Cross-Attention Fusion}
Table \ref{tab:ablation_sim}(B) validates that our proposed fusion strategy is critical for managing high-dimensional VFM priors. Reverting to MLP Concatenation incurs a $15.2\%$ absolute drop in success rate. This degradation suggests that unconstrained feature concatenation renders the policy susceptible to overfitting transient visual artifacts. Conversely, the Proprioceptive-Query Cross-Attention mechanism effectively modulates the visual focus toward terrain features conditioned on the instantaneous state. 
Furthermore, ablating the Proprioceptive History $\mathbf{h}_t$ severely diminishes the $R_{\mathrm{succ}}$ to $67.2\%$, highlighting the indispensability of temporal context for disambiguating partial visual occlusions induced by the robot's own limbs.

\subsubsection{Regularization as Physical Grounding}
The dual-head auxiliary estimator functions as an explicit regularization mechanism to inherently ground the high-dimensional latent space in physical geometry. Empirically, we observed that all auxiliary losses stably converge within 2,000 steps across diverse visual representations. Ablating the Terrain Reconstruction head induces a substantial performance decline, decreasing the $R_{\mathrm{succ}}$ from $86.4\%$ to $74.2\%$. This validates that auxiliary geometric supervision is imperative for preventing representation collapse and ensuring robust terrain traversal.
%%%%%%%%%%%%%%%%%%%%%%%%%%%%%%%%%%%%%%%%%%%%%%%%%%%%%%%%%%%%%%%%%%%%%%%%%%%%%%%%
\section{Conclusion}
\label{sec:conclusion}
We present \name{}, a purely RGB-driven reinforcement learning framework for humanoid locomotion. By conceptualizing monocular RGB as high-dimensional 3D latent representations via a frozen Visual Foundation Model (VFM), our approach achieves efficient visual-proprioceptive fusion through a lightweight cross-attention mechanism. Extensive simulations demonstrate that our dual-head auxiliary learning is crucial for explicitly grounding these visual priors in physical geometry. In zero-shot real-world deployments on the Unitree G1, the policy successfully negotiates challenging terrains such as stairs and low-light environments. These qualitative results validate the feasibility and robustness of VFM-derived features against visual degradation. Ultimately, \name{} provides a scalable paradigm that bridges geometric perception and semantic context, paving the way for next-generation Vision-Language-Action (VLA) agents.

\label{sec:conclusion}

\bibliographystyle{ieeetr}

\bibliography{glorified,new}

\end{document}